\documentclass[conference]{IEEEtran}
\IEEEoverridecommandlockouts
\usepackage{cite}
\usepackage{amsmath,amssymb,amsfonts}
\usepackage{algorithmic}
\usepackage{graphicx}
\usepackage{textcomp}
\usepackage{xcolor}
\usepackage{fancyhdr}
\usepackage{enumitem}
\usepackage{float}
\usepackage{multicol}
\usepackage{hyperref}
\usepackage{subcaption}
\usepackage{dblfloatfix}
\usepackage{algorithm}
\usepackage{algorithmic}
\usepackage{caption,subcaption}

\def\BibTeX{{\rm B\kern-.05em{\sc i\kern-.025em b}\kern-.08em
    T\kern-.1667em\lower.7ex\hbox{E}\kern-.125emX}}
\begin{document}

\title{Sketch2FullStack: Generating Skeleton Code of Full Stack Website and Application from Sketch using Deep Learning and Computer Vision
}

\author{Somoy Subandhu Barua\textsuperscript{1}, Imam Mohammad Zulkarnain\textsuperscript{1}, Abhishek Roy\textsuperscript{1}, Md. Golam Rabiul Alam\textsuperscript{1}, Md Zia Uddin\textsuperscript{2}\\ \textsuperscript{1}\textit{Department of Computer Science and Engineering, Brac University, Dhaka, Bangladesh} \\
\textsuperscript{2}\textit{Department of Sustainable Communication Technologies, SINTEF Digital, Oslo, Norway}
\\ \textit{Email: somoy.subandhu.barua@g.bracu.ac.bd, imam.mohammad.zulkarnain@g.bracu.ac.bd, } 
\\ \textit{abhishek.roy@g.bracu.ac.bd, rabiul.alam@bracu.ac.bd, zia.uddin@sintef.no}
}

\maketitle

\begin{abstract}
For a full-stack web or app development, it requires a software firm or more specifically a team of experienced developers to contribute a large portion of their time and resources to design the website and then convert it to code. As a result, the efficiency of the development team is significantly reduced when it comes to converting UI wireframes and database schemas into an actual working system. It would save valuable resources and fasten the overall workflow if the clients or developers can automate this process of converting the pre-made full-stack website design to get a partially working if not fully working code. In this paper, we present a novel approach of generating the skeleton code from sketched images using Deep Learning and Computer Vision approaches. The dataset for training are first-hand sketched images of low fidelity wireframes, database schemas and class diagrams. The approach consists of three parts. First, the front-end or UI elements detection and extraction from custom-made UI wireframes. Second, individual database table creation from schema designs and lastly, creating a class file from class diagrams.
\end{abstract}

\begin{IEEEkeywords}
deep learning, computer vision, full stack, EAST, tesseract OCR, UI wireframe
\end{IEEEkeywords}

\section{Introduction}
Since its creation, the internet needed to evolve to become more user-friendly and engaging and with that, server-based systems, such as full stack websites and applications, became more popular. The entire process involves developing the frontend from mockups and databases and classes from schemas and class diagrams respectively. Hence, this research addresses an automated way to tackle the growing demand of full stack websites from low fidelity UI wireframes and database tables and class diagrams.
 
Previously, there has been only a handful of studies which aimed to extract UI components from sketched images, but of those, a very small number trained their models with human-made sketches while the others mostly focused on training the models on actual wireframes, screenshots or automated images. Secondly, there is a scope of improvement for front-end UI detection as the stability and the performance can be further increased. Lastly, most of the research regarding the generation of code from sketches carried out earlier, worked on generating only the frontend portion of a full-stack application. Furthermore, as to our knowledge from extensive research there has not been any significant research on identifying and converting database schemas to create database tables as well as class diagrams to class files.
Based on the prior works carried out, we found a scope in research regarding the automation of the development of full-stack websites from sketches. As such, we proposed a novel approach in handling the main three components of a full-stack website or application which are the frontend, the backend and the database system. Back-end is the portion of the system that usually handles the business logics and connects the frontend through middleware or apis to the database~\cite{b1}. Whereas a database system comprises the database management system (DBMS), the software necessary for controlling it and the data itself~\cite{b2, b3}. In our research, we separately pursued generating the skeleton code or parsing of the frontend, database tables and classes from their respective diagrams.

This may not only significantly speed up the development process, saving both time and money, but also make the process of producing a full-stack application more accessible to a broader group of individuals in the long term. By providing an efficient deep learning and computer vision based approach, this research aims to generate the skeleton code of basic full stack website and application, especially frontend, database tables and classes from low-fidelity wireframes and associated diagrams. Therefore, our contributions in achieving this are as follows:

• We have detected and obtained information of frontend UI elements from manually drawn low-fidelity wireframe sketches. In order to perform that, we have created our own dataset of UI wireframe sketches using different tools to ensure diversification. Moreover, some custom designs for the elements were also introduced to further extend the representational capabilities of the UI wireframes. The images in the dataset were then passed through a series of preprocessing steps to remove unwanted noises before training the model for detection and classification of the elements from which we extracted the positional information and the type of UI element from similarly sketched UI wireframe images.

• We have generated queries for creating Database Tables from tool drawn Database Schemas. For executing this, we have applied Computer Vision techniques on individual table images of the database schemas where we identified the cells, then separated and extracted textual data from the images and processed all the data using an algorithm to generate the queries.   

• At the same time, we applied similar computer vision techniques on separate individual tables of tabular class diagrams to identify the fields along with their access modifiers within the class. The information was then processed through a custom made algorithm to generate the skeleton code of individual classes with their respective attributes, methods and access modifiers.

The generated classes can then be modified and used with frameworks to generate the model portion of the MVC (Model-View-Controller) framework. Moreover, by utilizing Persistence APIs like Java Persistence API (JPA) and repository interfaces such as CrudRepository with the modified generated code, it is possible to easily implement the functionalities of the CRUD (Create, Read, Update, Delete) operations within the codebase. This will ease the creation process of the api routes for CRUD operations hence the basic Controller logic as well.

\section{Related Work}

Nguyen et al.~\cite{b4}, tried to bridge the gap between the graphics designers involved in designing UI and the developers at the production process who are tasked with translating these concepts to working UI. REMAUI is the end result of their research. The purpose of REMAUI was to generate UI source code from application UI screenshots taken from Android and iOS systems. The methodology they employed to generate the expected result of source code was to use OCR to detect texts while Computer Vision techniques to detect contours of different elements based on hierarchies. With such an approach, they achieved good accuracy and were able to generate pixel-to-pixel similarity with the generated code. However, OCR based text recognition suffers from false-positives. Therefore, they had to rely heavily on domain specific knowledge and heuristics to differentiate between valid and non-valid words. For this case, they approached the problem with a computer vision technique called Canny’s Algorithm and then calculated the dilated edges’ contours. REMAUI then used the result to find atomic elements which were cross-checked with OCR results to figure out the actual text present in the screenshot. As such REMAUI was able to reject a significant number of false positives.  Though the generated source code demonstrates a somewhat complex nature when it comes to readability, it produced a near-accurate UI structure.

Another novel approach was proposed by  Beltramelli~\cite{b5} by the name of Pix2Code. The purpose of this research was to generate the UI source code from UI application screenshots from iOS, Android and web-based technologies. His methodology however, differed from the previous authors’ approach. They employed the combination of CNN for image feature extraction and Long Short-Term Memory (LSTM) for detecting Domain Specific Language (DSL) based tokens in order to generate code from screenshots.So, it can be inferred that Beltramelli’s dataset included the UI screenshots along with their respective DSL based context.  However, during the research, there was no predefined DSL to properly map the tokens with the UI components present in the screenshots, both of which were required to train the network. It thus relied on synthetic datasets which consisted of iOS, Android and web-based UI screenshots. As such it was able to generate good results, however, it is not optimized for real-world examples such as hand drawn wireframes with noises due to lighting and camera quality~\cite{b6}.

To tackle this and some other problems found in previous techniques, a different method has been proposed by Robinson~\cite{b7}. In his work, he automated the process of sketching the websites. The author normalized the website structures by removing all css, js and redundant html. Then, mapped his drawn sketch of components with the website codes to generate artificially sketched images. Here, the author worked only with input, image, button, paragraph and header tags (5 classes). As for the training methodology, instead of the Recurrent Neural Network and LSTM approach followed in Pix2Code, the author here approached the problem using two methods - classical computer vision and deep learning segmentation.  The classical computer vision approach consisted of the combination of Contour detection and Stroke-width transform (SWT) for text detection. The deep learning segmentation consisted of dilated convolutions via DeepLab v3+ with some key adjustments. After evaluating the two approaches it was evident that for most cases the deep learning segmentation outperformed the classical approach. However, for real hand-drawn sketches the first approach performed comparatively better. This suggested that a large number of hand-drawn sketches were required with a greater variety for the deep learning segmentation approach implemented by Robinson, to better generalize unseen sketching styles.

As previous related works and researches still had flaws while generating GUI code from hand-drawn real-world sketches, another improvisation was made in implementing sketch2code by Jain et al.~\cite{b8} In this research due to the emergence of various powerful deep learning techniques, the authors emphasized on converting hand-drawn sketches to UI real-time  using deep neural networks. The data-set contained 149 basic hand-drawn sketches with 2001 UI elements. The author did not apply any additional pre-processing to the input images aside from generating the boundary box coordinates and component classes into a CSV. For training the authors used RetinaNet with 50-Resnet layers pre-trained on ImageNet. The network also consisted of Feature Pyramid Network(FPN) for multi-scale feature maps that contain better quality information than the regular feature pyramid for object detection. For the second sub-problem regarding overlapping bounding boxes, the authors implemented a overlapping classes filtering algorithm that assigned individual priorities to the prediction classes and chose through priority-score hierarchy when overlap of two bounding boxes were over 50\%. Lastly, the third sub-problem which is to convert the predicted classes and bounding box into working UI code, a UI parser was designed which converted the UI representation object generated by the model into coded application.

A major component in full stack development is the back-end and the database. Extracting information from tables has been very challenging due to the variation in representation layouts. A multitude of approaches have been carried out in the past to tackle this problem. Gatos et al.~\cite{b9} used a set of predetermined rules to detect tables. The approach was based on the detection of vertical and horizontal lines and their point of intersection. These points of intersection are joined together with straight line segments to yield a reconstructed table. Although this approach beats the DSCC algorithm, it will fail to detect tables with rows and columns not bounded by line segments~\cite{b10}.

While most conventional table detection methods work based on the assumption that row-column boundaries of tables are straight and parallel, Seo et al.~\cite{b11} pointed out a problem with those approaches for table detection from camera captured images. Camera captured images may not always have straight and parallel row-column boundaries like scanned images from flatbed scanners due to curved document surface. To overcome this problem, his approach follows a distinct method for table detection from junctions and their labels. Camera captured images may have uneven illumination. To overcome this, first the captured image is converted to black-white image (binarization). Next, connected components (table boundaries, texts and noise) are extracted and through simple classification non tabular components are filtered out. To ease curve detection, a dynamic programming algorithm is used to break down the image into multiple regions where each region can have only one table. Now that each segment of the image has one table after table area detection, probabilistic progressive Hough transform (PPHT) detects locally straight-line segments that are extended on both ends. By applying the RANSAC algorithm repeatedly, the curvatures are refined. The intersection points of the horizontal and vertical curves are referred to as junctions. The junctions are then labelled from a set of 12 predefined labels. Along with the junctions and their labels, each cell can be located. This not only performed well in camera captured images but performed equally well in conventional scanned images.

However, lately deep learning demonstrated outstanding performance in image-based table detection~\cite{b12}. Gilani et al.~\cite{b10} presents an approach of detecting tables using deep learning. Faster RCNN (FRCNN) was opted for detection. Since FRCNN was initially proposed for classification of natural images, image transformation to separate text regions from non-text regions was necessary to train on existing models of FRCNN. The transformed image was then fed into Region Proposal Network (RPN) that generates rectangular bounding boxes around objects followed by a fully connected neural network for table detection. Through data augmentation from a very small dataset, in just 10000 iterations over momentum optimizer, this approach outperforms Tesseract~\cite{b13} by approximately 16\%.

Schreiber et al.~\cite{b14} brought forward a novel deep learning based solution not only for table detection but also would recognize the structure of the tables as two separate parts for born- digital PDFs and scanned documents. For table detection, the model is built by adopting RPN followed by FRCNN with ZFNet and VGG16 as the backbone. Using transfer learning technique in 1600 training images, in 30,000 iterations of 28 epochs with a batch size of 2 it achieved excellent accuracy on all three of its metrics. For structure recognition, FRCNN performed decently in column recognition while failing to detect rows with good performance. Instead, deep learning based FCN semantic segmentation was used. Unfortunately, the performance of the model was not comparable with other existing models due to being tested on randomly chosen ICDAR 2013 dataset and on raw images instead of PDFs.

Tesseract OCR~\cite{b15} is a novel approach in recognition of texts from images. It stores the outlines of the components and nest the outlines into Blobs. After which, the recognition is done in a two step process where the first step recognizes each word separately and the words with satisfactory probability are then passed into another adaptive classifier to learn from the existing words as well. Then another recognition is done in the entire document again to classify the words based on the newly learned parameters. In the final phase, fuzzy phases and the alternative hypotheses are checked to determine small cap texts~\cite{b16}.  For our research purpose, we proceeded with Google’s open-source codebase for Tesseract OCR v4.0. For the scope of our research, accurately predicting texts will be a crucial factor as mismatch in database attributes or class attributes in the skeleton code can be prone to high number of bugs during development. As such, among the three available options in this version we for all cases implemented the Tess-best module with Pytesseract which uses only LSTM unlike the other two options to detect texts more accurately yet slowly~\cite{b15}.

\section{Methodology}

In our research, the goal is to generate programmed interfaces from sketched images of not only the Frontend User Interface but also the Database system and the Backend code, we initially divided our work into three main segments with multiple sub-segments for each. Moreover, in each segment, we focused only on the particular portion of our Full-stack development process. The three main segments are - 

\begin{enumerate}[label=\roman*.]
    \item The Frontend UI Generation Segment
    \item The Database Generation Segment
    \item The Backend-Code Generation Segment
\end{enumerate}

As each segment poses different problems and subproblems, and as our approach to solving each is quite different compared to one another; we discussed each segment in separate subsections.

\subsection{The Frontend UI Generation}

To generate the programmed interface of UI elements from sketched images - first, we need to classify and localize all the UI components within the image. The image may also contain texts as part of the wireframe design which needs to be considered as well.  Therefore, we divided our Frontend UI Generation Segment into multiple sub-problems as well.

\begin{enumerate}[label=(\alph*)]
    \item First, we need to Detect and Classify the UI Elements in the images and understand their context.
    \item Second, we need to extract textual information from the images and generate their boundary boxes as well.
    \item The third problem consists of handling the bounding box overlaps in our first two steps. 
    \item The fourth and the final sub-problem is to convert the contexts generated from the wireframe sketches into an easily understandable Domain-Specific language (DSL).
\end{enumerate}

In this section,  we first describe our dataset and the preprocessing steps we have taken. We then explain our approaches in solving the above mentioned sub-problems in turns.

\begin{figure}[h]
\centering
\includegraphics[width = 0.9\columnwidth]{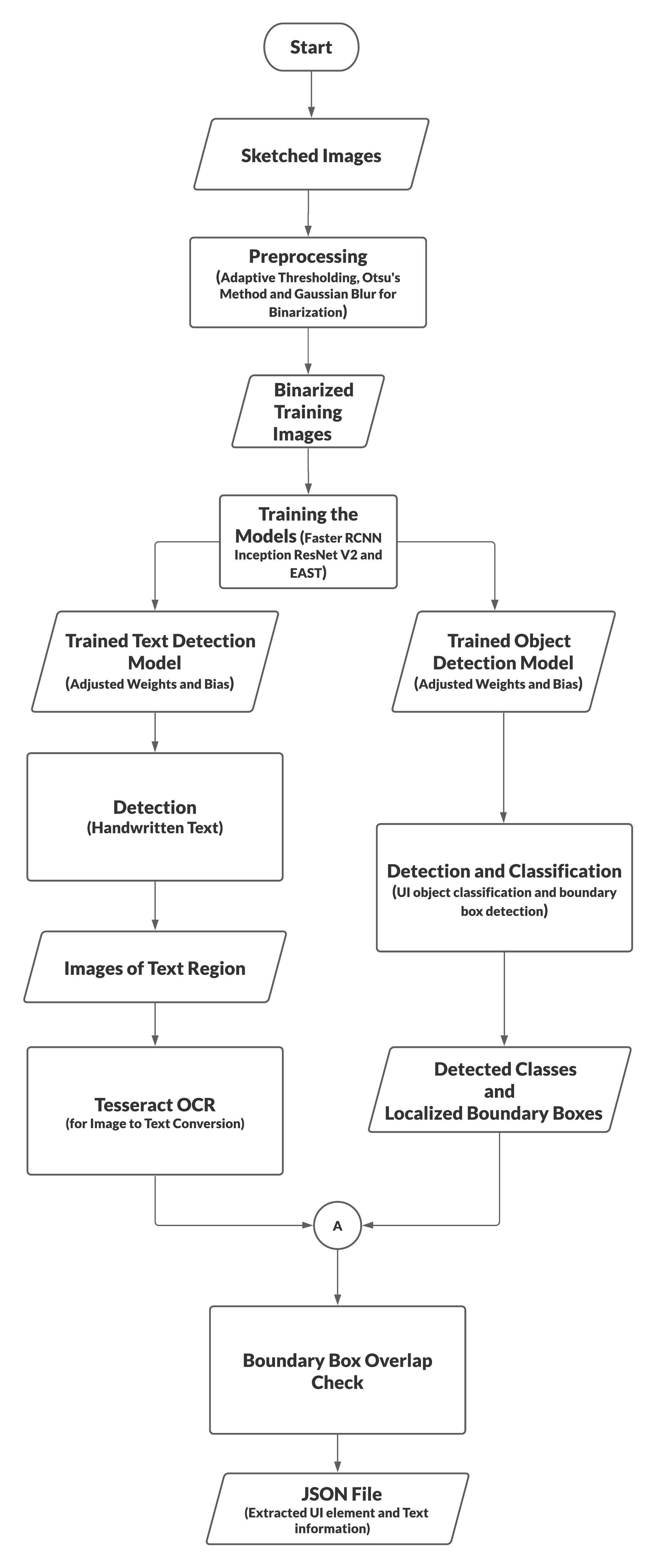}
\caption{Proposed Method for Frontend UI Generation}
\label{fig:x Proposed Method for Frontend UI Generation}
\end{figure}

The framework of our overall Frontend UI generation process is shown in Fig. ~\ref{fig:x Proposed Method for Frontend UI Generation}.

\subsubsection{Dataset}
For the front-end portion, our research focuses on generating UI elements from sketched images. This requires us to design the frontend mockups in such a way that it can be easily pre-processed and fed into our deep learning model. We also need to ensure that the abstractions from our image data and their contexts are easily learnable by our model. Therefore, we mostly followed the design rules of common wireframe symbols for UI elements but also added some of our own symbols for easier learning and better Full-stack integration.

However, adding new custom-made symbols for certain UI elements can cause confusion among different users, who might not be used to these newly designed UI wireframe symbols. Nevertheless, as the developers' final goal is to generate a fully working system with both frontend and backend where one frontend UI might be connected to other multiple frontend UIs, we deemed it as a necessity to have some alterations to the already existing commonly used symbols in order to obtain optimum results.

Due to having some changes to the traditional mockup designs and having our own set of symbols, we could not find any existing dataset of sketched frontend mockups that could fulfill our purpose as of date. As such, we manually sketched various mockups following our slightly modified wireframe design language.

\begin{figure}[h]
\centering
\includegraphics[width = 1\columnwidth]{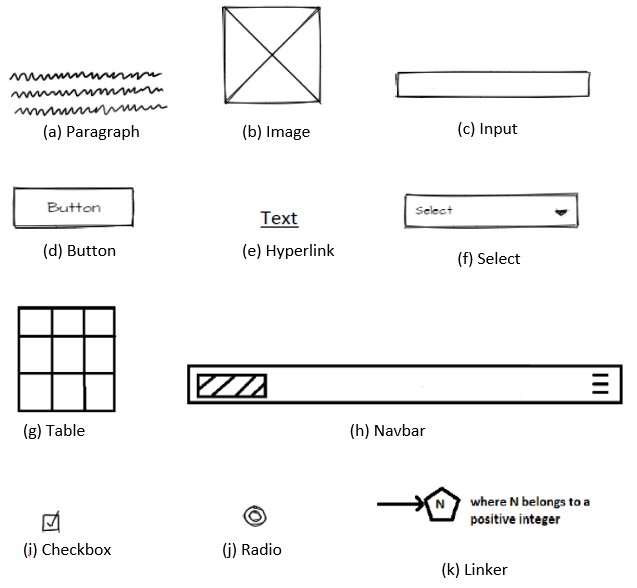}
\caption{Wireframe Classes} {(a) Paragraph (b) Image (c) Input (d) Button (e) Hyperlink (f) Select (g) Table (h) Navbar (i) Checkbox (j) Radio \\(k) Linker (for linking to other UIs)}
\label{fig:x Wireframe Classes}
\end{figure}

Our dataset for the frontend segment consists of a total of 11 classes, see Fig. ~\ref{fig:x Wireframe Classes}. Each class represents different low-fidelity wireframe symbols that we deemed most useful for generating the skeleton of frontend web pages and apps. 


\begin{figure*}[t]
    \centering
        \begin{subfigure}{0.18\textwidth}
        \centering
            \includegraphics[width=1\textwidth, height = 4cm]{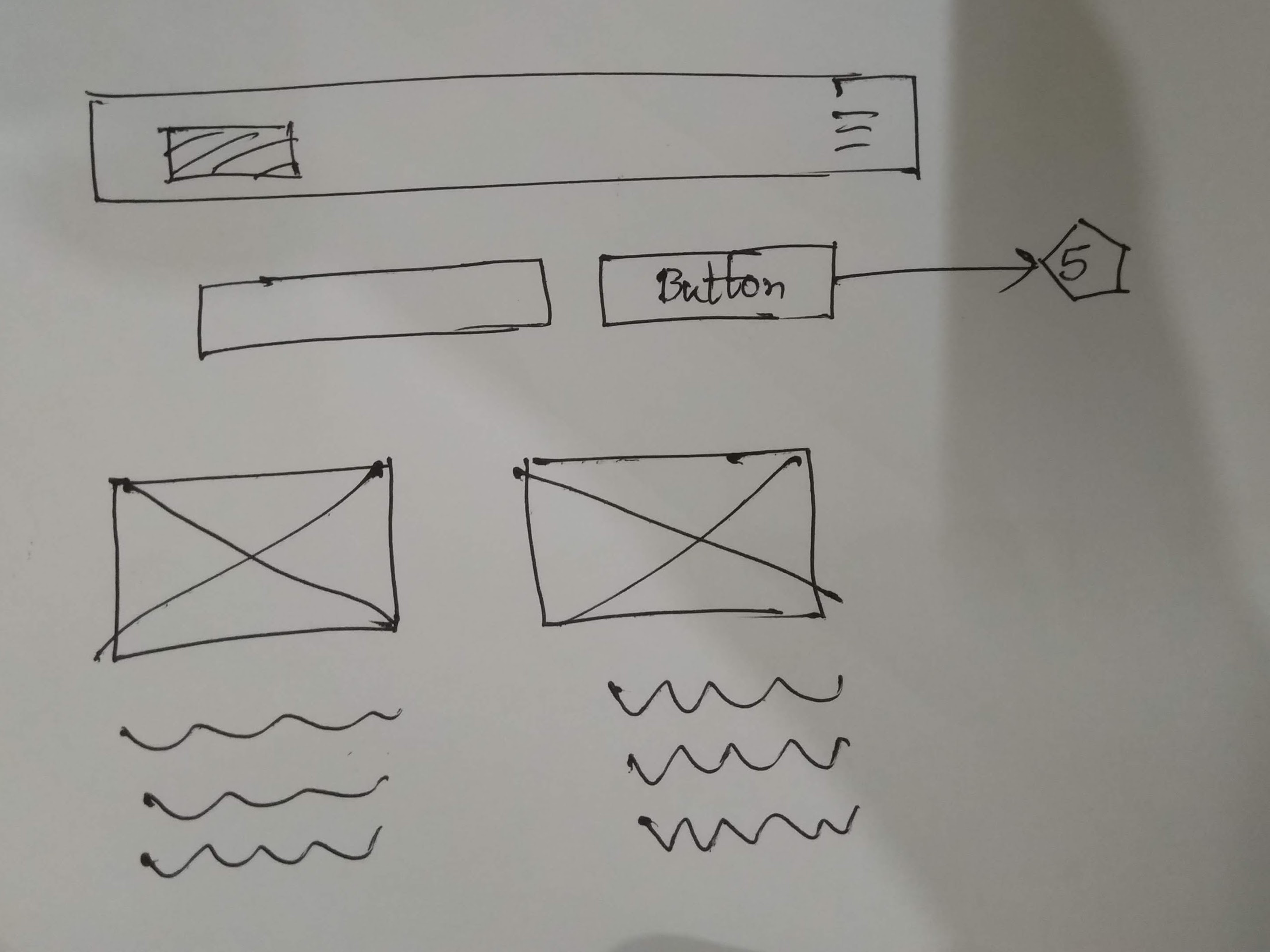}
            \caption{}
            \label{fig: Camera snapped image drawn with pen}
        \end{subfigure}
        \hspace{0.1em}
        \begin{subfigure}{0.18\textwidth}
        \centering
            \includegraphics[width=1\textwidth, height = 4cm]{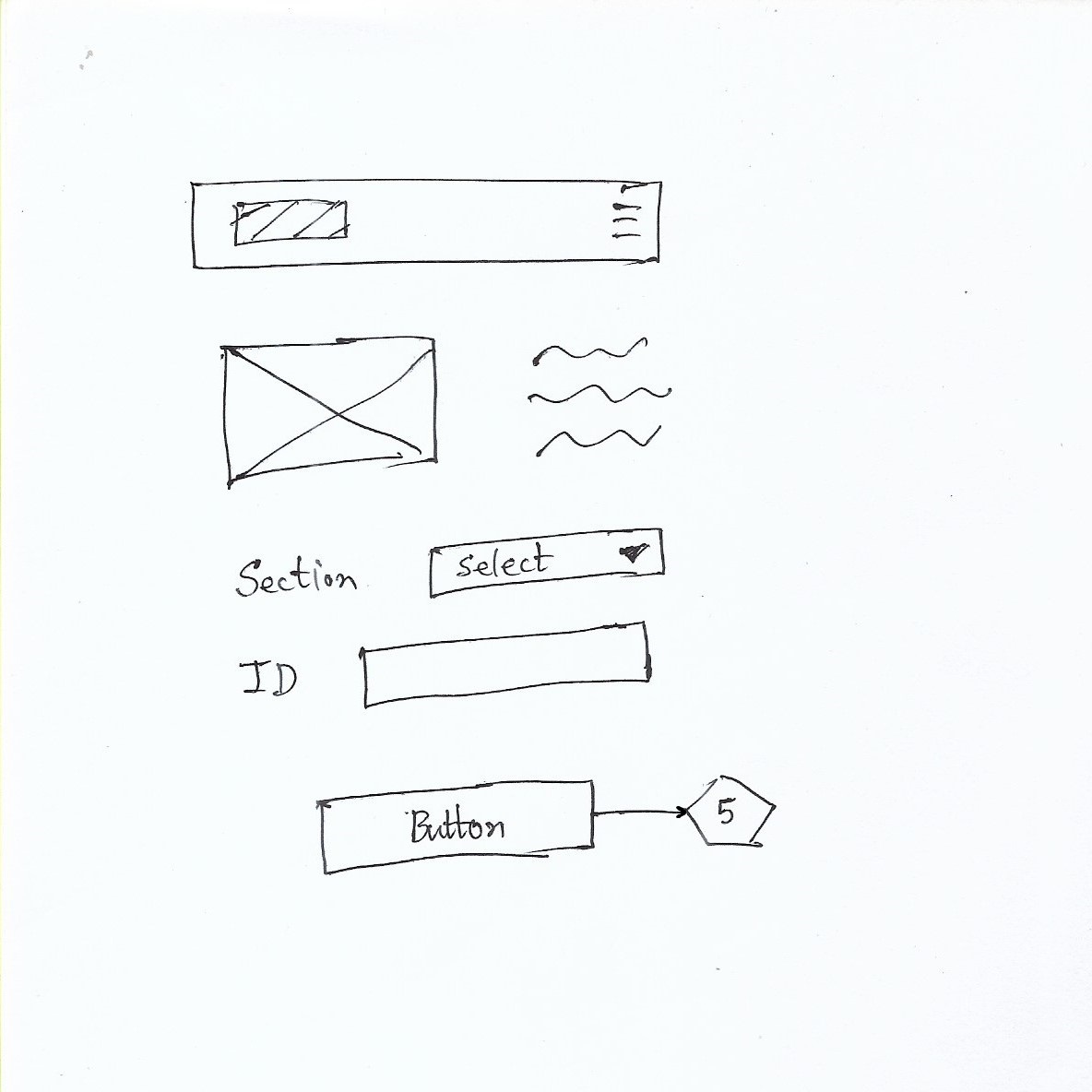}
            \caption{}
            \label{fig: Scanned image drawn with pen}
        \end{subfigure}
        \hspace{0.1em}
        \begin{subfigure}{0.18\textwidth}
        \centering
            \includegraphics[width=1\textwidth, height = 4cm]{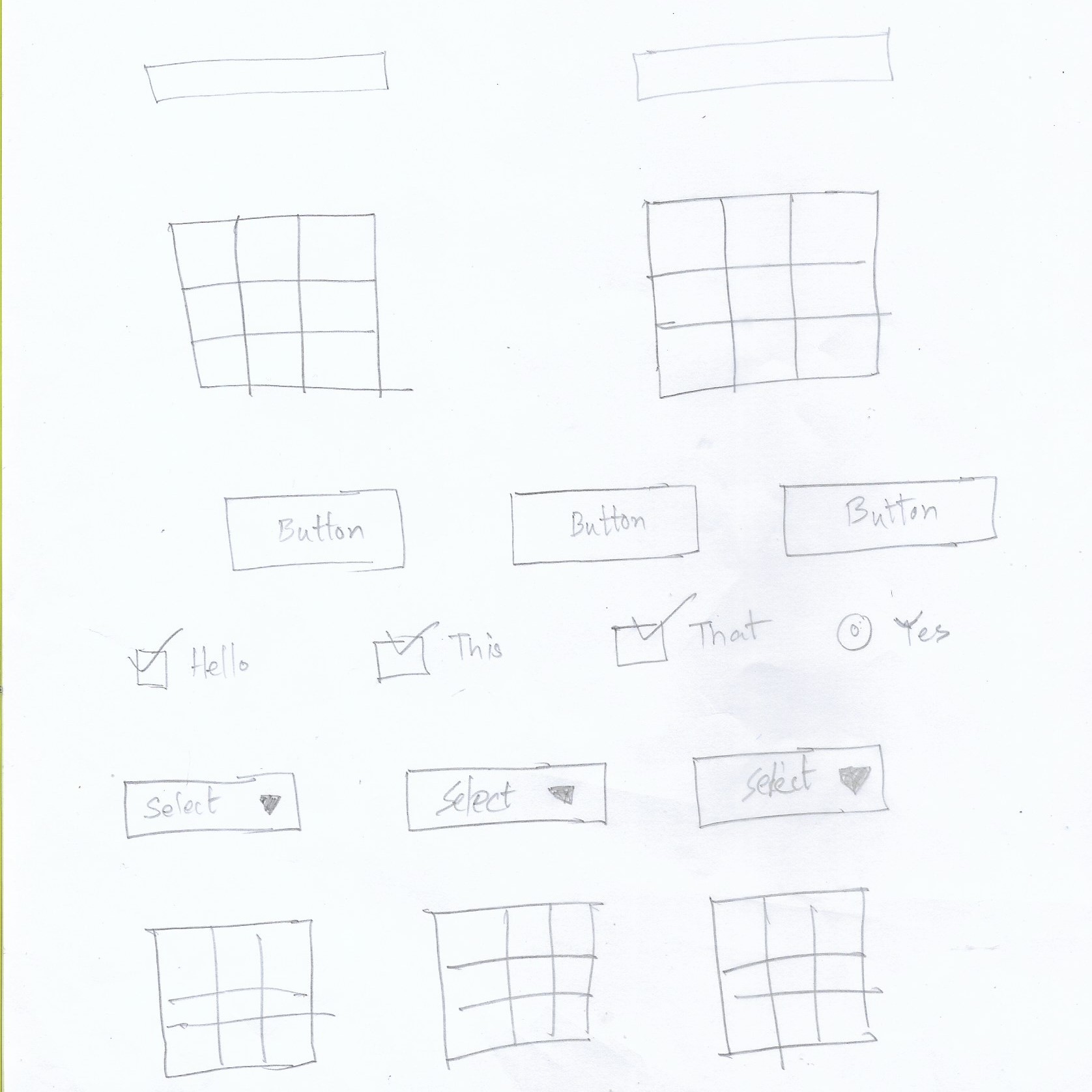}
            \caption{}
            \label{fig: Scanned image drawn with pencil}
        \end{subfigure}
        \hspace{0.1em}
        \begin{subfigure}{0.18\textwidth}
        \centering
            \includegraphics[width=1\textwidth, height = 4cm]{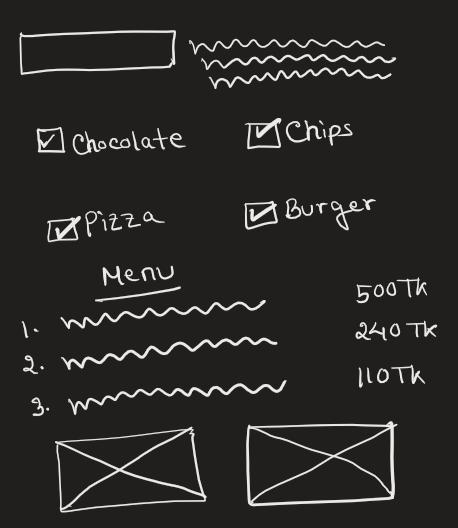}
            \caption{}
            \label{fig: Screen snipped drawn on black background}
        \end{subfigure}
        \hspace{0.1em}
        \begin{subfigure}{0.18\textwidth}
        \centering
            \includegraphics[width=1\textwidth, height = 4cm]{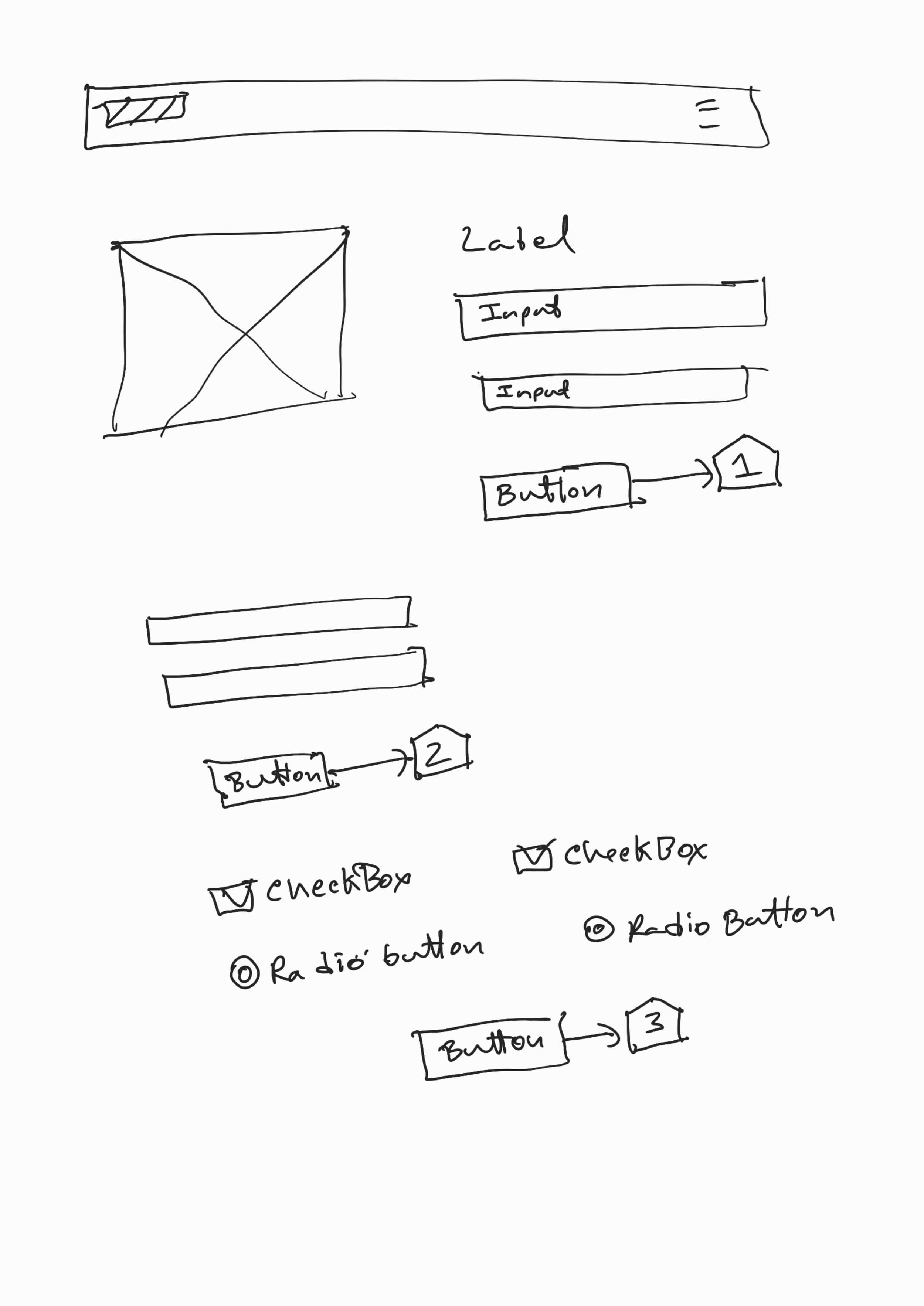}
            \caption{}
            \label{fig: Screen snipped drawn on white background}
        \end{subfigure}
    \caption{Different forms of images in the dataset} {(a) Camera snapped image drawn with pen (b) Scanned image drawn with pen (c) Scanned image drawn with pencil \\ (d) Screen snipped drawn on black background (e) Screen snipped drawn on black background}
    \label{fig:x Different forms of images in the dataset}
\end{figure*}


Among these classes, the Fig.~\ref{fig:x Wireframe Classes}(k) represents a ‘Linker’- class. To build a full-stack system, it is necessary to incorporate some action or event with UI elements. For example -  clicking on a button may trigger some event like posting data to the server or just run some simple javascript or even just redirect the current page to some other existing page. Thus, to incorporate these functionalities the ‘Linker’ class may come in handy. For our current scope of research, we will mainly focus on going to another page or running a script by utilizing the linker. The other mockup symbols shown above are very self explanatory. Hence, we will move on to our sketched images dataset.

Initially, we manually sketched wireframes following already existing websites. However, to add some variety to the image data, to consider various platforms used to draw wireframes nowadays, we used three different mediums to draw the mockup pages on.

\begin{enumerate}[label=(\alph*)]
\item We drew UI wireframes on pen and paper manually in the traditional ways. Then, for some images we took photos using different digital cameras under different lighting conditions.
\item We also scanned some of the hand-drawn sketches using a digital scanner.
\item To consider for tablet users, we also included sketched images manually drawn in iPad and Samsung Tablets.
\item Additionally, we also included sketches drawn with graphics tablets like Wacom and Huion.
\end{enumerate}

After taking all the above mentioned variety of images into consideration, we have made a total of 500 custom sketches with 5899 UI elements. We have mapped all the input images with the set of classes and labelled all the images with rectangular boundary boxes using LabelImg~\cite{b17} in PASCAL VOC format. The sketches and their respective labels were done by a group of people who either were well versed in program and web design or were simply new to website or application structure.

\subsubsection{Preprocessing}


\begin{figure*}[b]
    \centering 
\begin{subfigure}{0.15\textwidth}
  \includegraphics[width=1\textwidth, height = 3.5cm]{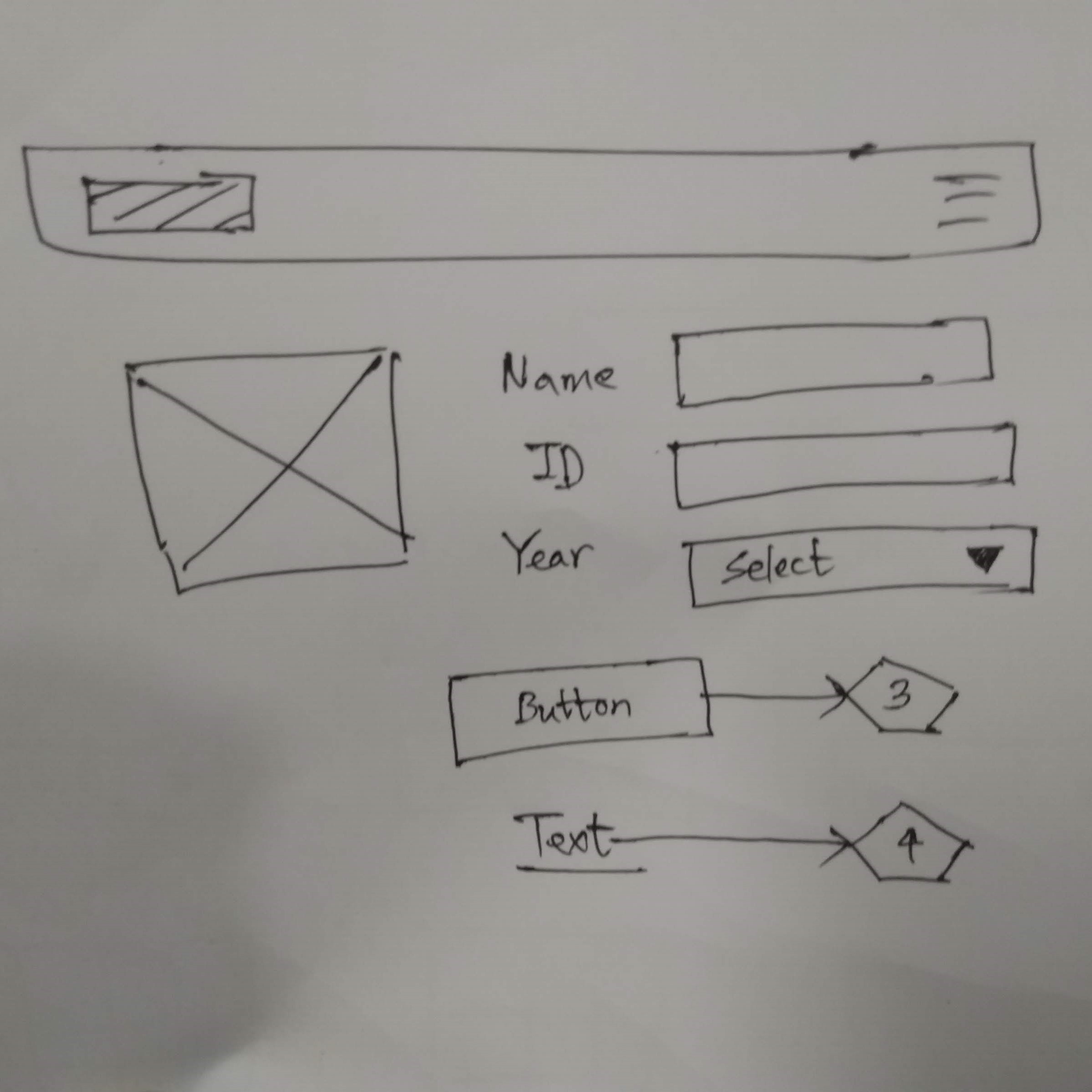}
  \caption{}
\end{subfigure}\hfil 
\begin{subfigure}{0.15\textwidth}
  \includegraphics[width=1\textwidth, height = 3.5cm]{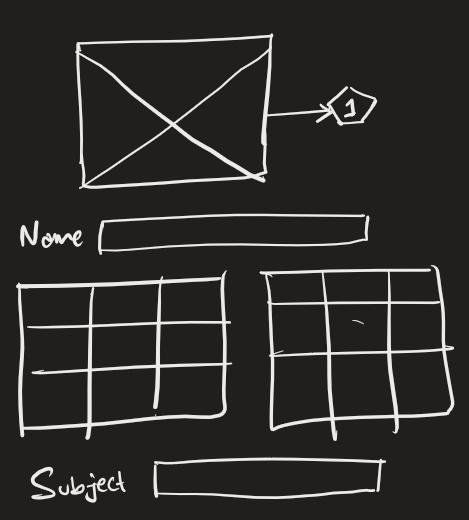}
  \caption{}
\end{subfigure}\hfil 
\begin{subfigure}{0.15\textwidth}
  \includegraphics[width=1\textwidth, height = 3.5cm]{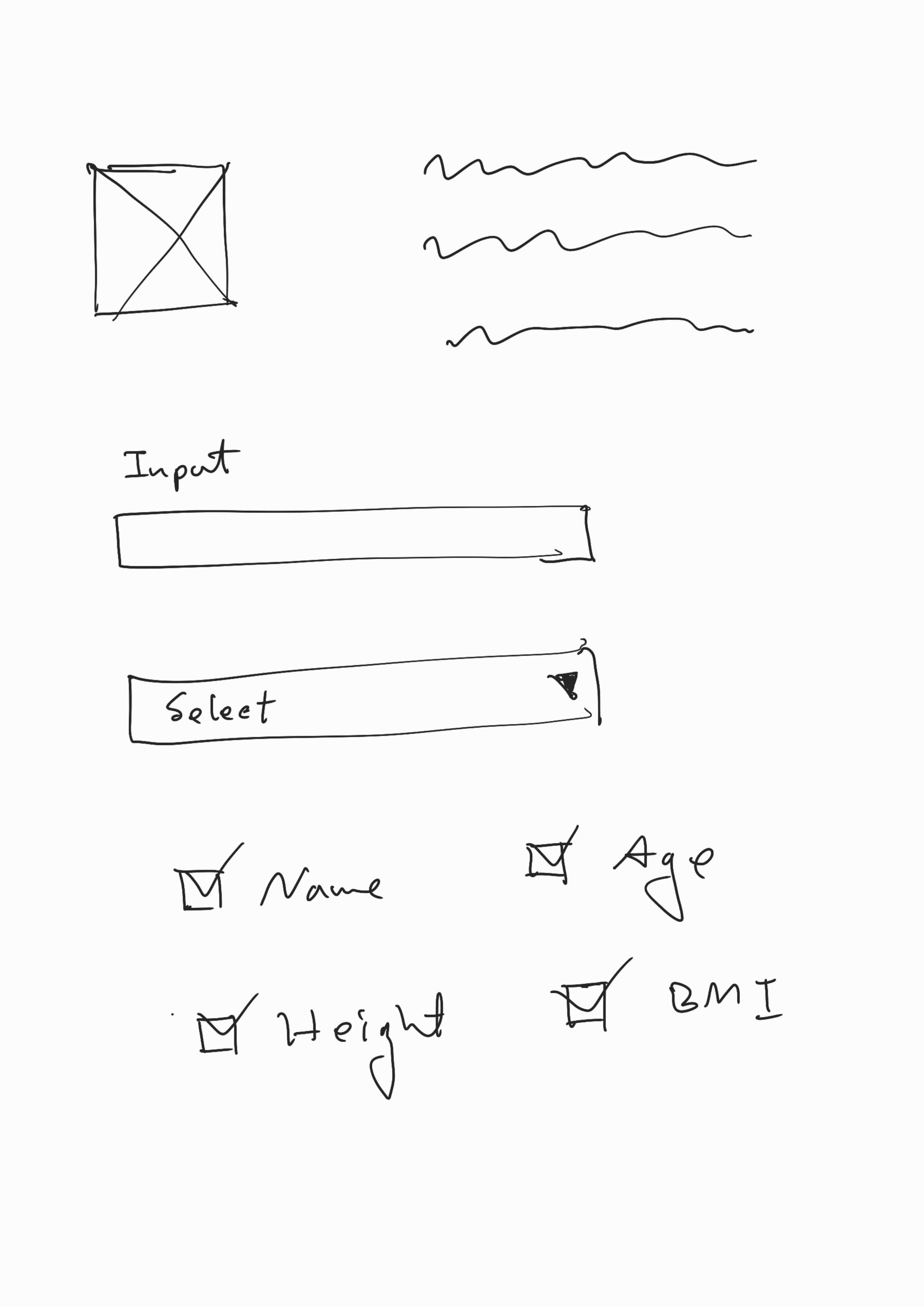}
  \caption{}
\end{subfigure}

\medskip
\begin{subfigure}{0.15\textwidth}
  \includegraphics[width=1\textwidth, height = 3.5cm]{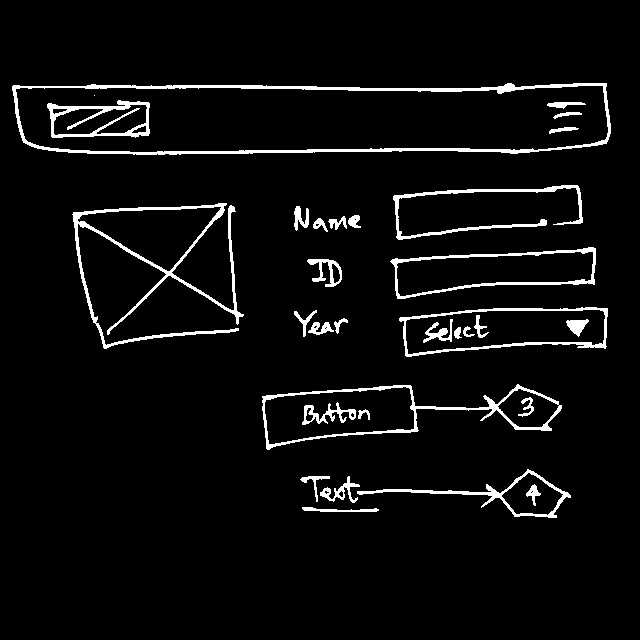}
  \caption{}
\end{subfigure}\hfil 
\begin{subfigure}{0.15\textwidth}
  \includegraphics[width=1\textwidth, height = 3.5cm]{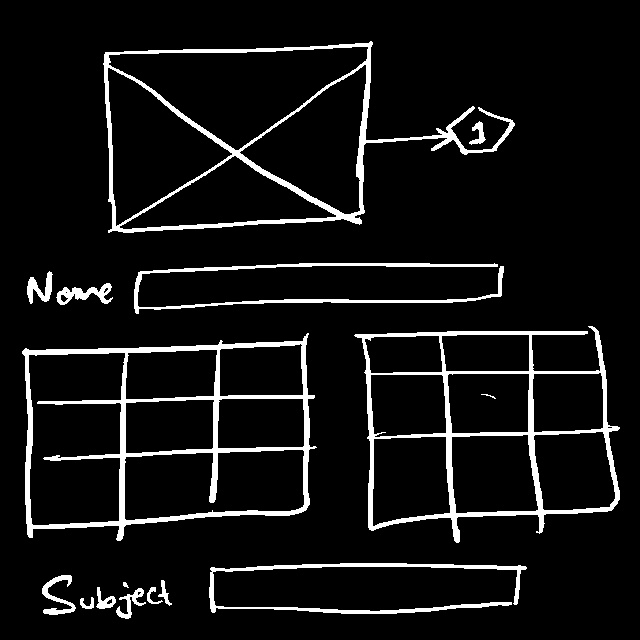}
  \caption{}
\end{subfigure}\hfil 
\begin{subfigure}{0.15\textwidth}
  \includegraphics[width=1\textwidth, height = 3.5cm]{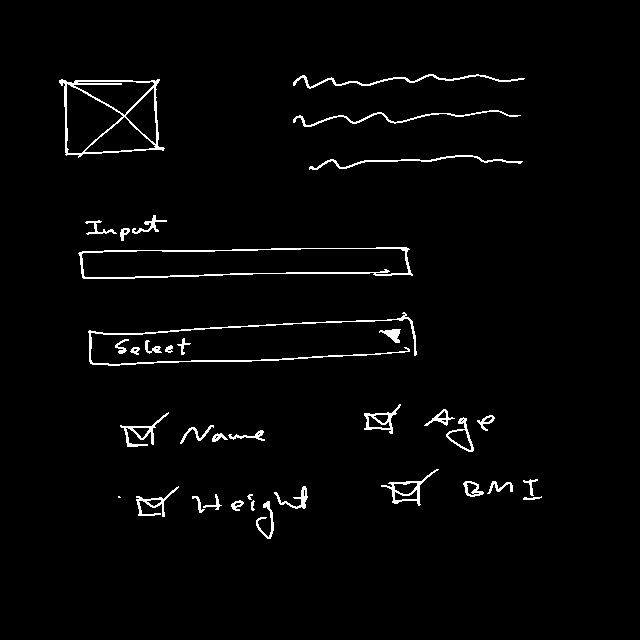}
  \caption{}
\end{subfigure}
\caption{Sketched image and their corresponding binarized form:} {(a), (b) and (c) are the sample input images before preprocessing; \\ (d), (e) and (f) are the respective binarized output images after preprocessing}
\label{fig:x Sketched image and their corresponding binarized form}
\end{figure*}


Unprocessed data fed into deep learning classification models, in most cases, do not produce high accuracy primarily due to noises that incorporate with it.  While the performance of deep learning models can be improved by tweaking certain layers and parameters, image preprocessing techniques can also hold utmost significance in boosting the accuracy~\cite{b18}. 

Before preprocessing, certain aspects were taken into consideration. Since we are only generating the skeleton code, color information of drawn UI elements was irrelevant. Hence we adopted binarization using thresholding to process the images. Binarization reduces the amount of information the image holds by separating pixels into two gray levels which not only removes the color information but also removes any unwanted pixels (noises) from the image. This also makes it viable to run in devices with low computational power and higher speed~\cite{b19}.

The gray level pixel values of the object in an image usually differ from that of its background. Thresholding capitalizes on this to extract the foreground of the image from the background~\cite{b20}.  Based on the category of thresholding (global or local) a certain threshold is passed. Global thresholding requires a single value of threshold (T) for binarization. The performance of global thresholding falls with degraded (noisy) images~\cite{b21}. On the other hand, Otsu’s method~\cite{b22} automatically computes the optimum threshold by calculating the weighted sum of intra class variance for all possible threshold values, provided the pixel intensities of the image have a bimodal distribution. The threshold value that gives the lowest weighted sum of intra class variance is passed as the threshold. The performance of Otsu’s method falls for noisy images which arises due to non-uniform lighting conditions.

To overcome the poor performance in varying illumination, adaptive thresholding on local windows is used. The key idea is that local regions have similar illumination. Thresholding over small regions with a certain region-wise value of T gives better performance over degraded images.

As we had different categories of images in our training dataset, different preprocessing techniques had to be used. For images drawn on white backgrounds (see Fig. ~\ref{fig:x Sketched image and their corresponding binarized form}(a) and ~\ref{fig:x Sketched image and their corresponding binarized form}(c)), mean adaptive thresholding was used over a 73x17 sized kernel. Since local thresholding is carried out on smaller regions instead of the entire image, selecting the kernel size is crucial. If the kernel size is too small, the foreground and the background pixels cannot be distinguished. So, it must be large enough to fit enough background and foreground images to carry out adaptive thresholding. However, if the kernel size is too large, the idea of similar illumination over smaller regions gets violated. For images drawn on dark backgrounds, adaptive thresholding did not perform too well. These images were drawn on Microsoft OneNote (see Fig. ~\ref{fig:x Sketched image and their corresponding binarized form}(b)). Since, the images were noise-free and had uniform illumination Otsu’s Method was used.

\subsubsection{UI Element Classification and Detection}
In the models discussed in Section-II, we can separate the state-of-the-art deep learning based object detection models into two categories. Multi-stage detectors and One-stage detectors.

Among the multi-stage detectors, we have two-stage detectors, such as Faster R-CNN (Region-based Convolutional Neural Networks)~\cite{b23} which 
\begin{enumerate}[label=\roman*.]
    \item Use a RPN (region-proposal network) to generate the regions of interest in the first stage.
    \item Then on the second stage, sends the regions through the model pipeline for classifying the objects and predicting their bounding box regressors.
\end{enumerate}

Two-stage models like Faster RCNN can achieve very high accuracy rates but usually take longer time to train.

On the other hand, single-stage detectors such as YOLO (You Only Look Once)~\cite{b24} treat both object classification and boundary box regression as a single regression approach by learning the class probabilities and bounding box coordinates from the input images together. Such models usually reach lower accuracy rates, but are much faster at detecting the objects in the image than two-stage object detectors~\cite{b25}.

In our research, we want to prioritize the accuracy of detecting the UI elements above all others. The aim is not to detect objects real-time like those in YOLOv3, YOLOv4 but to detect all the objects as close to ground truths as possible. Therefore, for this segment of our research we decided to choose Faster RCNN with Inception and Resnet V2 (640x640).

Point to be noted is that to compare how well our two-staged model performed against a state-of-the-art one staged detector in actual practical application - we trained our dataset in both Faster RCNN with Inception Resnet V2 and RetinaNet with Resnet101. However, in our trained model the one stage method- Retinanet despite being fast had a lot of missing ground truths or False Negatives. This further approved our decision in choosing the two-staged detector.


\begin{figure*}[ht]
  \centering
  \includegraphics[width=2\columnwidth]{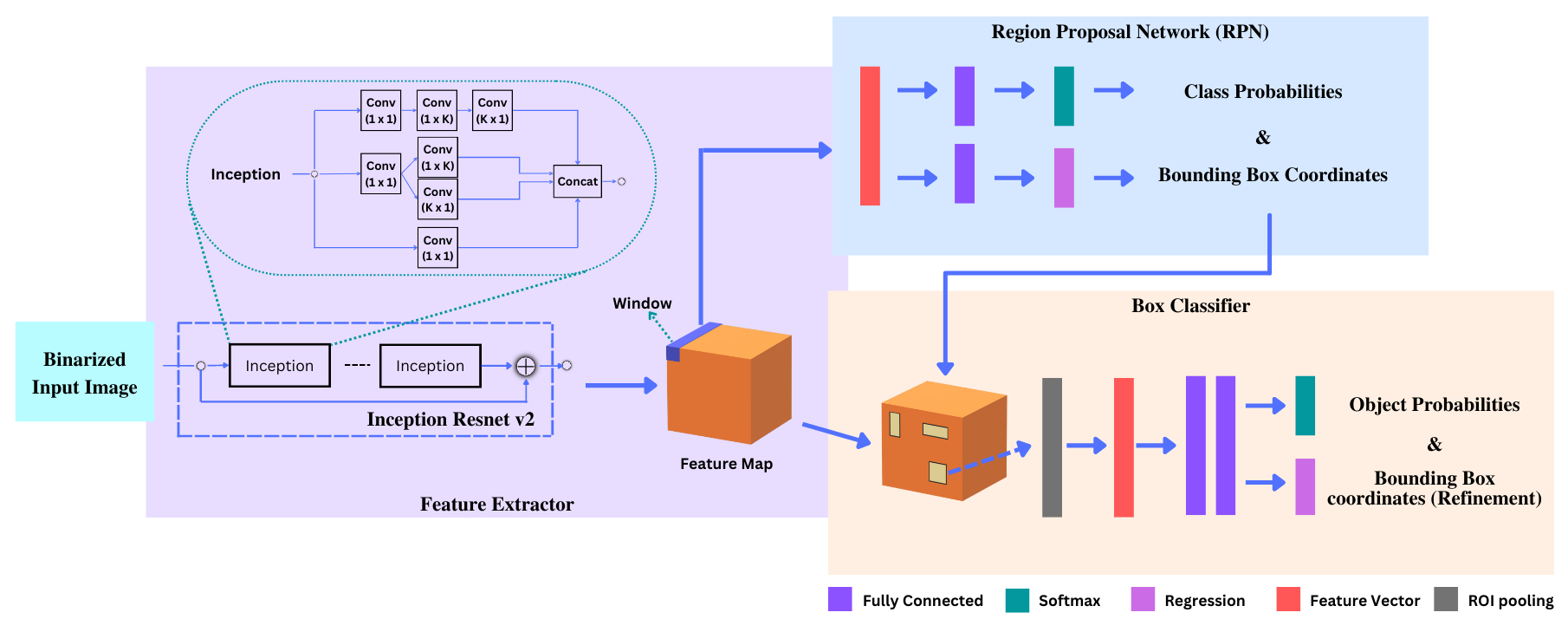}
  \caption{Meta architecture of Faster RCNN with Inception-Resnet-v2}
  \label{fig:x Faster RCNN}
\end{figure*}

Faster RCNN includes two stages as shown in Fig. ~\ref{fig:x Faster RCNN} ~\cite{b26},~\cite{b27},~\cite{b28},\cite{b29}. The convolutional feature map is generated by a feature extractor, Inception-Resnet-v2 in this case. The first stage of Faster RCNN is the Region Proposal Network (RPN) which takes a convolutional feature map as input.  A 3x3 spatial window is slid over the feature map. For each center point of the sliding window on the output feature map, 9 anchor boxes with 3 scales (128x128, 256x256, and 512x512) and 3 aspect ratios (1:1, 1:2, and 2:1) are used on the input image~\cite{b27}. Using two sibling fully connected layers, the proposed regions are determined - one layer (binary classifier) computes the probability if a proposed region contains an object (or belongs to a class) and the other layer (regressor) gives the coordinates of the bounding box~\cite{b28}. The training loss for RPN takes the equation as follows~\cite{b30}:

\begin{multline}
\label{eqn:1}
L(\{p_i\}, \{t_i\}) = \frac{1}{N_{cls}}\sum_iL_{cls}(p_i, p_i^*) + \\ \lambda\frac{1}{N_{reg}}\sum_ip_i^*L_{reg}(t_i, t_i^*)
\end{multline}

where in equation~\ref{eqn:1}, \(p_i\) denotes the probability of being an object for anchor with index i;
\(ti\), \(t_i^*\) denotes the coordinates of the predicted bounding box and the ground truth respectively.

\begin{equation}
\label{eqn:2}
L_{cls}(p_i, p_i^*) = -\log[p_i^*p_i+(1-p_i^*)(1-p_i)]
\end{equation}
where in equation~\ref{eqn:2}, \(L_{cls}(p_i, p_i^*)\) denotes the cross entropy loss on whether the object is present or not

\begin{equation}
\label{eqn:3}
L_{reg}(t_i, t_i^*) = R(t_i-t_i^*)
\end{equation}
and in equation~\ref{eqn:3}, \(L{reg}(t_i, t_i^*)\) denotes the difference between the regression of predicted box and the ground truth where R is the smooth L1 function

\begin{equation}
\label{eqn:4}
\emph{smoothL1}(x) = \begin{cases}
                        0.5x^2 \times \frac{1}{\sigma^2} & |x| < \frac{1}{\sigma^2} \\
                        |x| - 0.5 & \text{otherwise} \\
                    \end{cases}
\end{equation}

In the second stage, Fast RCNN is used. Based on the box proposals, regions of interest (ROI) are generated from the feature map which is later fed to the pooling layer and fully connected layers to compute the object probability and refined bounding box coordinates.

The Faster RCNN with Inception-resnet v2 model we used is a pre-trained model that was  released by Google. It has been trained on the COCO dataset and is completely open-source. The model was built on top of Tensorflow, with the intention of easily constructing, training, and deploying object detection models. The reason we selected this specific model repository is because of its past research on the support of both object detection and localization tasks~\cite{b31},~\cite{b32}.

We trained the model with our input images and their mapped classes in Pascal VOC format for a total of 40,000 steps. As discussed previously, our custom dataset was trained using pre-trained weights obtained from the Faster\_rcnn\_inception\_resnet\_v2\_atrous\_coco~\cite{b31}.

We kept the hyper-parameters the same with the values provided in the pipeline.config of the model repository.  The experiment was performed using the framework of Tensorflow object detection API on a Windows machine with a Nvidia GeForce GTX 1070 MAX-Q with 8GB GDDR5 dedicated GPU memory and additionally 16GB DDR4 RAM.

\subsubsection{Text Detection and Extraction}
To detect text in the sketched images, EAST is used.  EAST~\cite{b33} is a two staged text detector that is faster than traditional text detection approaches. It uses a fully convolution network that predicts the bounding boxes of the text regions. The predictions are passed to Non-Maximum Suppression (NMS) that retains the bounding box with the highest confidence score. The model can be changed to either predict words or lines. Even though EAST is a robust model that can detect text at an average of 13.6fps, its implication in our case is simple and straightforward.  The RGB sketched images, resized to 640x640, instead of the preprocessed binarized images are fed into the EAST model, trained on the ICDAR 2013 \& 2015 dataset, which carries out the localization of text in the sketched UI wireframe design.

Our objective is to convert the texts from the images into strings such that it can be directly mapped in our application. However, all the texts are not required to be converted to strings. For instance, the text "Button" and “Select” (refer to table x), is a signature for the button and dropdown UI element, respectively. Hence, should not be converted unlike the texts that come along with radio buttons and the checkboxes.

The area of overlap between the bounding box of EAST(BBE) and the bounding box of UI element (BBU) over the area of BBE, Intersection over EAST(IoE), is computed. By comparing IoE with a threshold of 0.5, we determine if the text is required to be converted to string. Algorithm~\ref{alg: Retain EAST bounding box for conversion to string} is run on all images to retain the bounding box of EAST that is subsequently passed through tesseract OCR for image to string conversion.

\begin{equation}
    \text{Intersection over EAST (IoE)} = \frac{\text{Area of Overlap}}{\text{Area of EAST bounding box}}
\end{equation}

\begin{algorithm}
    \caption{Retain EAST bounding box for conversion to string}
    \label{alg: Retain EAST bounding box for conversion to string}
    \begin{algorithmic} 
        \FOR{each \emph{EAST bounding box}}
            \STATE $flag \leftarrow false$
            \FORALL{\emph{UI element bounding box}}
                \STATE \textbf{compute} IoE
                \IF{IoE $>$ 0.5}
                    \STATE $flag \leftarrow true$
                    \STATE $break$
                \ENDIF
            \ENDFOR
            \IF{$flag$ == $false$}
                \STATE \textbf{retained} \emph{EAST bounding box}
            \ENDIF
        \ENDFOR
    \end{algorithmic}
\end{algorithm}

  \begin{figure*}[b]
    \centering
        \begin{subfigure}{0.18\textwidth}
        \centering
            \includegraphics[width=1\textwidth, height = 4cm]{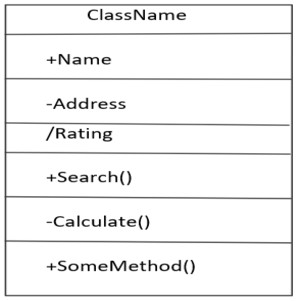}
            \caption{}
            \label{fig: Input Image}
        \end{subfigure}
        \hspace{5em}
        \begin{subfigure}{0.18\textwidth}
        \centering
            \includegraphics[width=1\textwidth, height = 4cm]{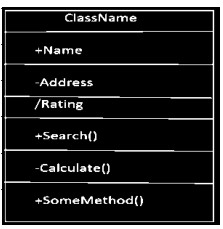}
            \caption{}
            \label{fig: Binarized Image}
        \end{subfigure}
        \hfill
        \begin{subfigure}{0.18\textwidth}
        \centering
            \includegraphics[width=1\textwidth, height = 4cm]{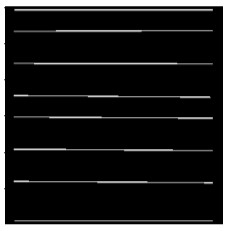}
            \caption{}
            \label{fig: Horizontal Lines Detected}
        \end{subfigure}
        \hfill
        \begin{subfigure}{0.18\textwidth}
        \centering
            \includegraphics[width=1\textwidth, height = 4cm]{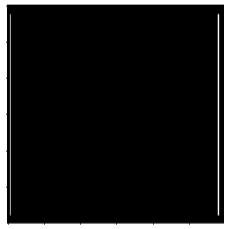}
            \caption{}
            \label{fig: Vertical Lines Detected}
        \end{subfigure}
    \caption{Steps of table image pre-processing:} {(a) Input Image (b) Binarized Image (c) Horizontal Lines Detected (d) Vertical Lines Detected}
    \label{fig: Steps of table image processing}
\end{figure*}

\subsection{Database and Class Generation}
In this research, as the main purpose is not to generate a fully built full-stack application but rather the skeleton of it from low-fidelity wireframes, schemas and design, we proposed solving the Database Table generation from database schemas and Class generation from class diagrams in the following way:-
\begin{itemize}
    \item As the backend of a full-stack application can be quite complicated with varying implementations and is often large in scale therefore, as discussed in Chapter-2, we opted for class generation from class diagrams. Diagrams such as class diagrams are often used to represent different code blocks of the system. For our research, we focused on class diagrams even though there exists different diagrams for representing different aspects of the system because, by generating the basic code structure of classes from class diagrams we can create the meta-structure or the skeleton of the code-base which developers can later easily enhance based on their own requirements and constraints - which is the target of our research. 
    \item For the database portion, we similarly focused on simple database schemas to generate the SQL code for the given database tables. However, generation of database table following our algorithm is not only limited to SQL but other query languages as well.
    \item In this research, we generated the skeleton code for the individual tables and the classes from the diagrams. Generating the code for the connection between database tables and classes can be complicated and may lead to various issues including security issues and entire system failures with low accuracy or false prediction. As such, the approaches to securely generate the ‘connections’ between each class or table has been left for future research.
\end{itemize}

To elaborate on the methodologies followed in our research, for generating database codes in structured query languages like MySQL from sketched images of database schemas and for generating class-skeleton codes from class diagrams, detecting all the database tables in the schema can be a necessary process. Database tables in the schema are quite similar to ordinary tables with mostly a single column and multiple rows. As such, table detection approaches such as CascadeTabNet~\cite{b34}, CDEC-Net~\cite{b35} , TableNet~\cite{b36} can play a significant role in detecting such tables.

For the type of tables from database schemas  and class diagrams that we are working in this research, the tables from the diagrams can be easily detected using TableNet~\cite{b36}. However, for the text extraction phase, tables from class diagrams and database schemas contain a single column with multiple rows separated by horizontal lines. It is important to note TableNet is not specialized for exactly this purpose and rather the model is generalized and is designed to tackle a broad set of tables with no horizontal lines separating the rows for its data extraction. 
Therefore, to specialize for our specific use case, we followed a different approach to extract the information of a single table (both class table and database table) and then generate its respective code skeleton.

The images of each class diagram are individually processed to detect and read the contents of each row. Both global and Otsu’s thresholding~\cite{b22} is performed on the images for binarization. On the binary image a morphological operation is carried out using rectangular kernels. Following Marius’s approach in his article~\cite{b37}, for the extraction of vertical lines, the kernel (vertical) is given a dimension of Nx1 and for horizontal lines the kernel (horizontal) of 1xN is used, where N = 1/50th the width of the image. Erosion is performed on the image separately, once using the vertical kernel for vertical line extraction and next the horizontal kernel for horizontal line extraction. Erosion retains a pixel value as 1 if all the other pixels under the kernel is 1.

On the eroded horizontal lines, dilation is performed using horizontal kernels. Dilation sets a pixel value to 1 if any of the pixels under the kernel is 1. The same is done on the eroded vertical lines using the vertical kernel defined before. The eroded lines are made more prominent after dilation. The image with the horizontal lines and the image with the vertical lines are merged into a single image with equal weights, 0.5, to get the tabular structure. The resulting image is eroded once more with a rectangular 2x2 kernel and then thresholded.

Unlike Marius’s approach~\cite{b37}, this research makes some modification necessary to get better cell detection. Contouring joins all the continuous points with the same intensity. In order to fetch each of the contour's bounding box coordinates all of the contours are first needed to be identified. The mean of the heights of the bounding boxes are computed by using those contours whose bounding box heights are within a reasonable range relative to the height of the input image which in turn will be essential for identifying rows and columns. This approach is taken to prevent invalid contours, for instance, the whole table as a single contour is not considered as a valid cell. As a result, for determining row and columns, the subsequent y coordinate of the bounding box is compared with the previous box. If the y coordinate of a box is smaller than the y coordinate of the previous box by half the mean height, the box is considered a column, otherwise stored as rows. The coordinates of the bounding boxes are stored in a list. From each region of interest, the noise is removed via further morphological operation before Tesseract OCR is run over to convert the text from image to string.

\subsubsection{Class Generation}
Upon using the Tesseract OCR~\cite{b15} on the image after processing, a list of strings for the detected texts is created and stored in detected\_texts variable. This list is passed to a function known as class\_generation (see Algorithm~\ref{alg: class generation}). There are mainly three main parts to a class that this function focuses on and as such are assigned with three variables. The class\_name is initialized with the first element from the detected texts list, the class\_attributes is set to an empty list, the same with class\_methods. Next, for every remaining element from the detected texts list, the access modifier is searched and set and depending on the presence of “(“ and ”)”, that particular element along with its access modifier is appended to their respective list class\_attributes or class\_methods. Now, a single character prefix determines the access modifier of the methods and the properties/attributes of the class. A “+” symbol indicates public, a “-” symbol indicates private and a “/” means it is protected. If no such symbols are present then it can be considered a default access modifier which based on the programming language that it is going to be translated into can be public, private or protected. At the end, this function can call another function which can be used to fill up a class template with values from class\_name, class\_methods and class\_attributes. For this research, a function named class\_file\_maker was called with these variables as parameters which created a python class file. Algorithm~\ref{alg: class generation} outlines this process.

  \begin{algorithm}
    \caption{function \emph{class\_generation()}}
    \label{alg: class generation}
    \begin{algorithmic}[1]
        \STATE \textbf{initialize} a variable \textbf{class\_name} and empty lists \textbf{class\_methods}, \textbf{class\_attributes}
        \STATE $class\_name \leftarrow detected\_text[0]$ 
        \FOR{remaining \textbf{element} in detected\_text} 
            \IF{"-" found in \textbf{element}}
                \STATE \textbf{remove} "-" from \textbf{element}
                \STATE $access\_modifier \leftarrow "private"$
            \ELSIF{"/" found in \textbf{element}}
                \STATE \textbf{remove} "/" from \textbf{element}
                \STATE $access\_modifier \leftarrow "protected"$
            \ELSIF{"+" or no access modifier found in \textbf{element}}
                \STATE \textbf{remove} "+" from \textbf{element} if it exists
                \STATE $access\_modifier \leftarrow "public"$
            \ENDIF
            \IF{"()" found in \textbf{element}}
                \STATE \textbf{append} [element, access\_modifier] to class\_methods
            \ELSE
                \STATE \textbf{append} [element, access\_modifier] to class\_attributes
            \ENDIF
        \ENDFOR
        \STATE \textbf{call} class\_file\_maker(class\_name, class\_attributes, class\_methods)
    \end{algorithmic}
\end{algorithm}

\subsubsection{SQL Query Generation from Database table}
The Tesseract OCR~\cite{b15} when used on the processed image’s region of interest for the database table image a list named detected\_text was generated. This list is passed to the  database\_generation (see Algorithm~\ref{alg: database generation}) as a parameter.This then checks for the existence of primary keys and stores them in a variable called primary\_keys. Additionally, for each element in the detected\_text, the particular attribute of the table and its corresponding data type are also stored in another variable table\_attributes. Later, these two variables are passed to another function called create\_sql\_query which basically outputs the table creation SQL command to create the table as per the input image. 

  \begin{algorithm}
    \caption{function \emph{database\_generation(detected\_text, table\_name)}}
    \label{alg: database generation}
    \begin{algorithmic}
        \STATE \textbf{initialize} empty lists \text{primary\_keys}, \textbf{table\_attributes}
        \FOR{each \textbf{element} in detected\_text}
            \STATE split \textbf{element} 
            \STATE $attribute \leftarrow element[0]$
            \STATE $data\_type \leftarrow element[1]$ 
            \IF{“*” found in \textbf{attribute} }
                \STATE \text{remove} "*" from \textbf{attribute}
                \STATE \textbf{append} attribute to primary\_keys
            \ENDIF
            \STATE \textbf{append} [attribute, data\_type] to table\_attribute
        \ENDFOR
        \STATE \textbf{call} create\_sql\_query(table\_name,  table\_attributes, primary\_keys)
    \end{algorithmic}
  \end{algorithm}

\section{Implementation and Results}
As discussed before in our training section, we trained our model with 500 hand-drawn binarized sketched wireframe  images. The training with the proposed model was conducted on a system with the following specifications:

\begin{table}[H]
\centering
\begin{tabular}{ | c | c |}
\hline
CPU  & Intel Core i7 8750H \\
\hline
GPU & Nvidia GeForce GTX 1070 Max-Q\\ & with 8GB GDDR5 memory\\
\hline
RAM & 16 GB DDR4 \\ 
\hline
Tensorflow & 2.5 \\ 
\hline
Python & 3.8 \\
\hline
\end{tabular}
\caption{System Specification}
\label{tab: System Specification}
\end{table}

The training was set for 40,000 steps. Once the training was complete, results were readily available to us through Tensorflow’s Tensorboard.

After training, mAP@.50 which means that the IoU threshold for TP with the ground truth was 50\% or more gave us a score of 0.9944. While  mAP@.75 meaning the IoU threshold was 75\% gave us a score of 0.8845. Remember that mAP is the average of the area under the curve of the Precision-Recall graph for each category.

Second, the average Recall score. After training, the AR@100 was 0.8323. This means that if an image has 100 such elements and as a result 100 ground truths, then the model will successfully detect around 83 of those elements.

Third, the total loss. This encompasses bounding box classification and localization loss and the Region Proposal Network’s (RPN)~\cite{b38} localization loss and objectness loss along with regularization loss. The overall shape of the total loss graph is a downward curve meaning that throughout training it is able to achieve less and less loss. Therefore, after 40,000 steps, the total loss at 40,000 step was 0.07172 for the training phase and 0.2488 for the testing phase.

\begin{table}[H]
\centering
\begin{tabular}{ | c | c | }
\hline
Total Loss (Train) & 0.07172 \\
\hline
Total Loss (Eval) & 0.2488 \\
\hline
mAP@.75IOU & 0.8845 \\ 
\hline
mAP@.50IOU & 0.9944 \\ 
\hline
AR@100 & 0.8323 \\
\hline
\end{tabular}
\caption{Results}
\label{tab: Results}
\end{table}

\begin{figure}[H]
    \centering
    \begin{subfigure}{0.22\textwidth}
        \centering
        \includegraphics[width=1\textwidth, height = 4cm]{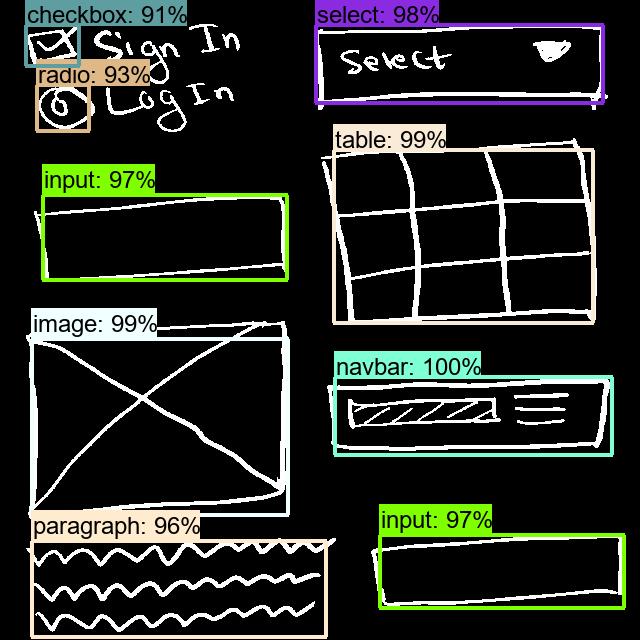}
    \end{subfigure}
    \hfill
    \begin{subfigure}{0.22\textwidth}
        \centering
        \includegraphics[width=1\textwidth, height = 4cm]{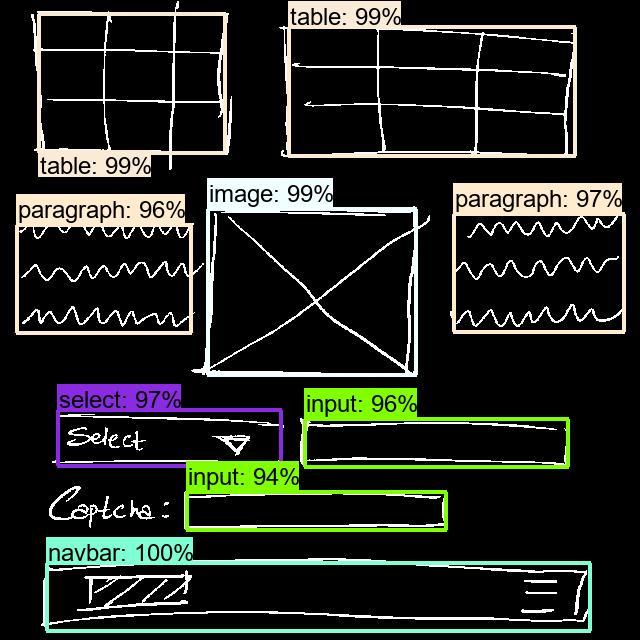}
    \end{subfigure}
    \caption{Inference Image of Object Detection}
    \label{fig: inference image}
\end{figure}

Lastly, the overall result of the trained model. The two inference images (see Fig. ~\ref{fig: inference image}) show how well the model performs when given samples of test image for detecting and classifying.

From Fig. ~\ref{fig: inference image}, it can be seen that it is able to detect all of the elements present in the two images and is able to classify them correctly with high confidence scores.

For the text detection and extraction sub-problem, we implemented EAST: An Efficient and Accurate Scene Text Detector~\cite{b33},~\cite{b39}. The model of EAST that we implemented for our dissertation has been pre-trained using the ICDAR 2013 and ICDAR 2015 training sets. Below are two inference images using this model.

\begin{figure}[H]
    \centering
    \begin{subfigure}{0.22\textwidth}
        \centering
        \includegraphics[width=1\textwidth, height = 4cm]{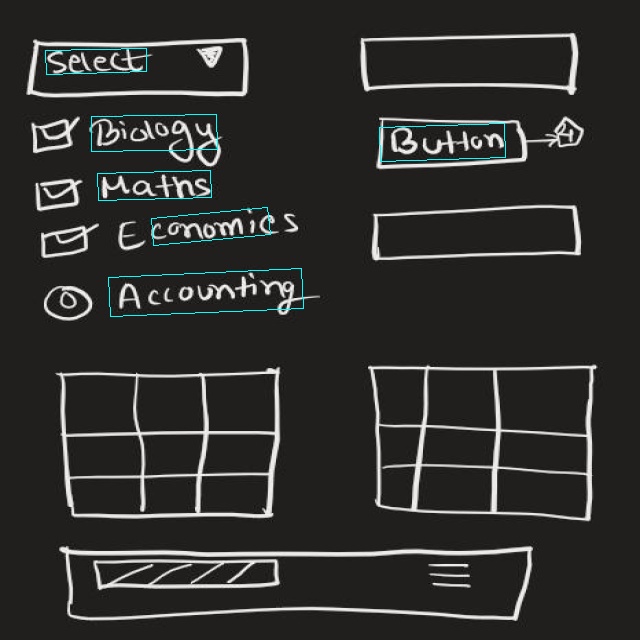}
    \end{subfigure}
    \hfill
    \begin{subfigure}{0.22\textwidth}
        \centering
        \includegraphics[width=1\textwidth, height = 4cm]{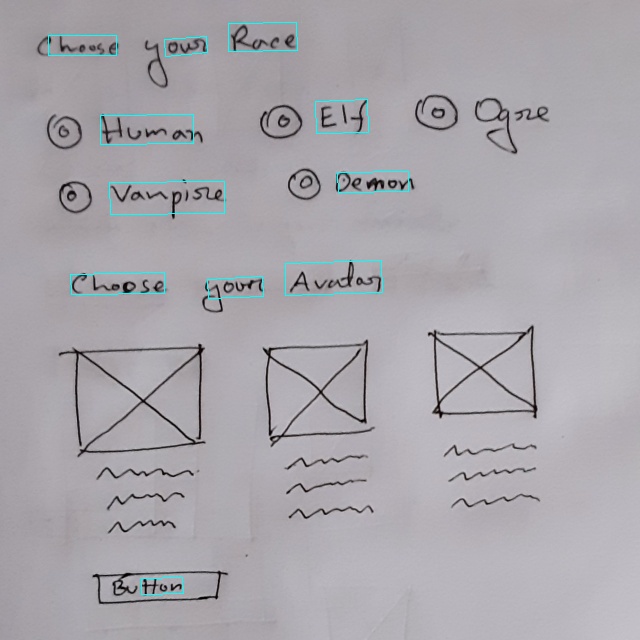}
    \end{subfigure}
    \caption{Inference Image of Text Detection}
    \label{fig: Inference Image of Text Detection}
\end{figure}

\begin{figure}[h]
    \centering
    \includegraphics[width=8cm]{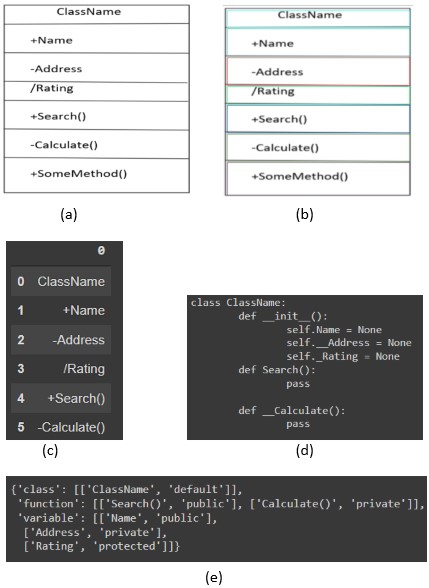}
    \caption{Steps of class generation.} {(a) Tool Drawn Input Image of Class Diagram (b) Cells Detected (c) Tesseract OCR converted text to string (d) Code for class generation (e) Extracted attributes and functions from class diagram }
\label{fig: Steps of class Generation.}
\end{figure}

\begin{figure}[h]
    \centering
    \includegraphics[width = 1\columnwidth]{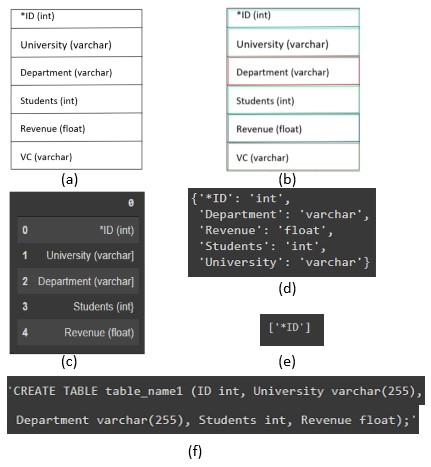}
    \caption{Steps of Database table creation.}{(a) Tool Drawn Input Image of Database table (b) Cells Detected (c) Tesseract OCR converted text to string\\ (d) Extracted data from database table (e) Primary Key\\ (f) SQL query for given database table}
\label{fig:x Steps of Database table creation.}
\end{figure}

Fig. ~\ref{fig: Inference Image of Text Detection} shows EAST was successful enough to detect most of the natural texts from given test images as shown in the cyan colored bounding boxes. Our approach will then use the text boundary box coordinates and the UI element boundary box coordinates and will determine whether there is an overlap between the two. This is done by checking if the coordinates of the UI element envelops the coordinates of the text by a certain threshold set by us. If such a case exists, then it will not be sent for text extraction using OCR based text extraction which in our case is Pytesseract~\cite{b40}. For instance, for the above image on the left, the last element is a “Button”. It is a rectangle with the word “Button” in it. As it is evident from the image, EAST has detected that particular word. So now our approach would be to get its coordinates and compare it with the coordinates of the UI element “Button”. It will be seen that UI element coordinates are outside the coordinates of the text. Thus, it will be disregarded as an element of text.

From there, we can use JSON to create an object file consisting of the necessary information required to convert it to any output format.

After the frontend generation from sketched images, we generated python code (.py) files from tables from class diagram containing the attributes and properties of the class in the class diagram table has been shown in Fig. ~\ref{fig: Steps of class Generation.}.

For the database portion (Fig. ~\ref{fig:x Steps of Database table creation.}), the SQL query that is generated from the given database table was on par with what was expected with the input image. Additionally, if needed, the primary key(s) of the databases can be separately stored as well. This allows the database engineers and developers to understand whether the primary keys are accurately recognized and easily connect different database components with one another after the initial skeleton is generated.

\section{Conclusion and Future Work}
The process of automating the development of full stack websites and applications from their designed schemas can be complicated. Although our research tries to tackle the generation of skeleton codes for three important parts of a full stack application- the front-end, classes , and the database tables; our research has its own limitations as well. For instance, additional css, animation and features can be integrated into the front-end portion with high-fidelity wireframes whereas our research mostly focuses on low-fidelity wireframes. Additionally, in this research database tables and class files were generated from database schema and class diagrams. However, generating full database systems and connected class components is still an opportunity for future work.

Nevertheless, with the advancement in technology, the overgrowing demand of full stack website development and application is on the rise. IT firms need to invest ample resources in the design phase of such websites and apps. Keeping those designs in mind and then writing the codes is a major part of their workflow which can be quite time consuming. As such, generating the skeleton codes from sketches can really ease up the overall process. If this research can save even the least amount of time in the development process and ease the workload on the developers and if it would make full stack development a lot more accessible to a greater audience and can be considered as a significant contribution in the field.

\end{document}